\documentclass{article} 
\pdfoutput=1
\usepackage{nips14submit_e,times}
\usepackage{hyperref}
\usepackage{url}
\usepackage{graphicx}
\usepackage{amsmath,amsfonts,amssymb}
\usepackage{soul}

\definecolor{lightgreen}{rgb}{.5,1,0}
\sethlcolor{lightgreen}
\newcommand{\highlight}[1]{\hl{\textbf{#1}}}

\title{Modelling, Visualising and Summarising Documents with a Single Convolutional Neural Network}

\author{
Misha Denil$^1$ \quad Alban Demiraj$^1$ \quad Nal Kalchbrenner$^1$ \\
\textbf{Phil Blunsom}$^1$ \quad
\textbf{Nando de Freitas}$^{1,2}$ \\
$^1$University of Oxford, United Kingdom\\
$^2$Canadian Institute for Advanced Research\\
\texttt{\{misha.denil,nal.kalchbrenner,phil.blunsom,nando\}@cs.ox.ac.uk}\\
\texttt{a.demiraj@oxfordalumni.org}}

\nipsfinalcopy 

\newcommand{\vS}{\mathbf{S}}
\newcommand{\vw}{\mathbf{w}}
\newcommand{\vW}{\mathbf{W}}
\newcommand{\vf}{\mathbf{f}}
\newcommand{\vF}{\mathbf{F}}

\newcommand{\R}{\mathbb{R}}

\begin{document}

\maketitle

\begin{abstract}
Capturing the compositional process which maps the meaning of words to that of documents is a central challenge for researchers in Natural Language Processing and Information Retrieval. 
We introduce a model that is able to represent the meaning of documents by embedding them in a low dimensional vector space, while preserving distinctions of word and sentence order crucial for capturing nuanced semantics.
Our model is based on an extended Dynamic Convolution Neural Network, which learns convolution filters at both the sentence and document level, hierarchically learning to capture and compose low level lexical features into high level semantic concepts. 
We demonstrate the effectiveness of this model on a range of document modelling tasks, achieving strong results with no feature engineering and with a more compact model.
Inspired by recent advances in visualising deep convolution networks for computer vision, we present a novel visualisation technique for our document networks which not only provides insight into their learning process, but also can be interpreted to produce a compelling automatic summarisation system for texts.
\end{abstract}


\section{Introduction}

Encoding symbolic concepts with distributed representations is a dream that has excited researchers for decades~\cite{Hinton1986,Bengio2003}.  In recent years this idea has re-emerged with the successes of neural networks in language modelling~\cite{Mikolov2013a}, machine translation~\cite{Schwenk2012,Kalchbrenner2013a,Devlin2014} and other natural language processing tasks such as chunking and named entity recognition~\cite{Collobert2011}.  

A particularly fruitful vein of recent research has tackled compositional models of vector space semantics. 
While algebraic approaches have proved popular for their simplicity \cite{Mitchell2010,Grefenstette2011}, it is, arguably, approaches based on deep neural networks, which have generated the most recent interest. This is exemplified by work on recursive neural networks that have been used to great effect for embedding sentences for tasks such as sentiment analysis~\cite{Socher2011,Socher2012,Hermann2013a}.

Most work fusing NLP and neural networks has been based upon fully connected networks, in either an entirely feed forward manner~\cite{Socher2012} or including recurrent connections~\cite{Sutskever2011,Kalchbrenner2013a,Kalchbrenner2013}.  A notable exception to this is the model of Kalchbrenner~\emph{et~al.}~\cite{NalKalchbrennerGrefenstette2014}, which uses a convolutional neural network (ConvNet for short) to build continuous distributed representations for sentences.

A fundamental building block of nearly all applications of neural networks to NLP is the creation of continuous representations for words.  Since text is fundamentally symbolic and neural networks operate on continuous inputs, it is necessary to create a mapping from the symbolic representation of text into the continuous space in which the networks operate.  The creation of word embeddings has received much attention, and several excellent methods now exist for creating these mappings. In addition to providing a suitable representation for neural networks, word embeddings have been shown to capture many semantic relationships between the concepts they represent.  Word embeddings learned by neural networks also serve as an excellent general purpose representation for words that can be used in non-neural models.

In this paper we show how ConvNets can be used to build distributed representations of documents.  Our model is compositional; it combines word embeddings into sentence embeddings and then further combines the sentence embeddings into document embeddings.  The combinations at every level use ConvNets inspired by the convolution networks that have seen great success in computer vision.  Since our model is based on convolutions, it is able to preserve ordering information between words in a sentence and between sentences in a document. This information is lost in bag-of-words or n-gram models.

Going beyond classification, we also show how visualisation techniques developed in the computer vision literature for understanding the activation patterns of ConvNets~\cite{Zeiler2012,Simonyan2013}, can be applied directly to our convolutional document model.  These visualisations give direct insights into what the models learn. Furthermore, they can also be used to identify important sections of a sentence or document.  As a novel application of this we show how to use the visualisation technique of Simonyan~\emph{et~al.}~\cite{Simonyan2013} to automatically generate summaries for movie reviews.


\section{Model description}

Our model is divided into two levels, a sentence level and a document level, both of which are implemented using ConvNets.  At the sentence level we use a ConvNet to transform embeddings for the words in each sentence into an embedding for the entire sentence.  At the document level we use another ConvNet to transform sentence embeddings from the first level into a single embedding vector that represents the entire document.

The model is trained by feeding the document embeddings from the second level of the model into a softmax classifier, and the ConvNets in both levels are trained jointly by backpropogation through the entire model.  At the sentence level the weights between ConvNets that process different sentences are tied, so every sentence in the document has an embedding which is produced by the same ConvNet.

The levels in our model work at different levels of abstraction: the first level operates on words in a sentence and the second level operates on the sentences in a document.  The transformations in each level are performed by ConvNets; each ConvNet contains one or more layers of convolution, pooling and tanh transformations.

The architecture of our model forces information to pass through an intermediate sentence based representation.  This architecture is inspired by Gulechere and Bengio~\cite{Gulcehre2013} who show that learning appropriate intermediate representations helps generalisation, and also by Hinton~\emph{et~al.}~\cite{Hinton2011} who show that by forcing information to pass through carefully chosen bottlenecks it is possible to control the types of intermediate representations that are learned.

For both levels in our model we use a modified version of the Dynamic Convolutional Neural Network (DCNN) from Kalchbrenner \emph{et~al.}~\cite{NalKalchbrennerGrefenstette2014}.  This model is similar to the convolutional networks used in computer vision with a cascade of  convolution, pooling and tanh transformations, but has been adapted for text modelling.

\begin{figure}
\includegraphics[width=1.0\linewidth]{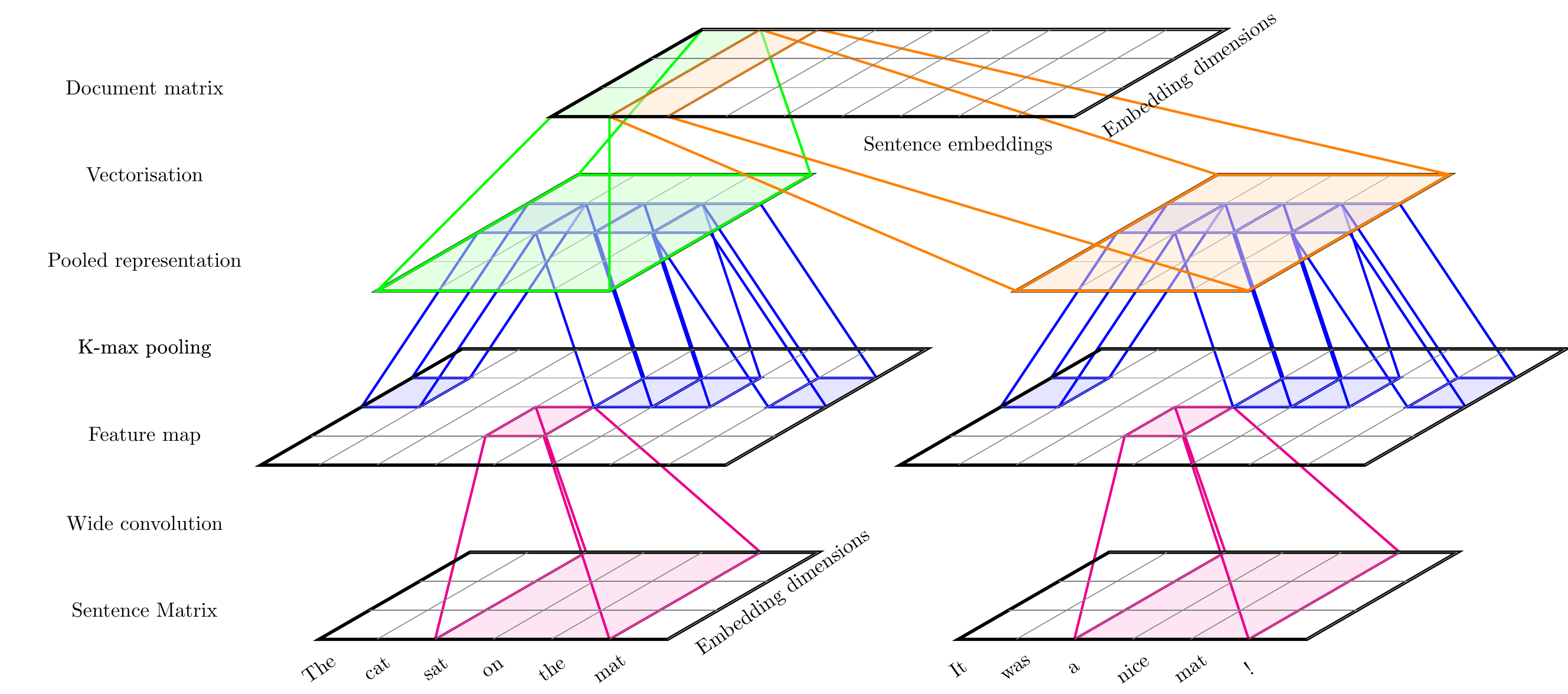}
\caption{Word embeddings are concatenated into columns to form a sentence matrix.  The ConvNet applies a cascade of convolution, pooling and nonlinearity operations to transform the projected sentence matrix into an embedding for the sentence.  The sentence embeddings are then concatenated into columns to form a document matrix.  The document model then applies its own cascade of convolution, pooling and nonlinearity operations to form an embedding for the whole document which is then fed into a softmax classifier.}
\label{fig:cdm}
\end{figure}

A detailed schematic of the sentence level of our model is shown in Figure~\ref{fig:cdm}, and an overview of the full model is shown in Figure~\ref{fig:compare-conv-style} (right).  In the following sections we describe how the convolutional layers operate within each level of our model.

\subsection{Embedding matrix}

The input to each level of our model is an embedding matrix.  At the sentence level the columns of this matrix correspond to embeddings of the words in the sentence being processed, while at the document level the columns correspond to sentence embeddings produced by the sentence level of the model.

At the sentence level, the embedding matrix is built by concatenating embeddings for each word into the columns of a matrix.  The words are drawn from a fixed vocabulary $V$, which we represent using a matrix of word embeddings $\vW \in \R^{d\times |V|}$. Each column of this matrix is a $d$ dimensional vector that gives an embedding for a single word in the vocabulary.  The word embedding vectors are parameters of the model, and are optimised using backpropogation.

\begin{align*}
\vW = \begin{bmatrix}
| & | & | \\
\vw_{1} &  \cdots & \vw_{|V|} \\
| & | & |
\end{bmatrix}
\end{align*}

For each sentence $s = \begin{bmatrix} w^s_1 & \cdots & w^{s}_{|s|} \end{bmatrix}$ we generate a sentence matrix $\vS_s \in \R^{d\times|s|}$ by concatenating together the word embedding vector corresponding to each word in the sentence.  

\begin{align*}
\vS_s = \begin{bmatrix}
| & | & | \\
\vw_{w^s_1} &  \cdots & \vw_{w^s_{|s|}} \\
| & | & |
\end{bmatrix}
\end{align*}

The sentence level of the model produces an embedding vector for each sentence in the document.  The input to the document level is obtained by assembling these sentence embeddings into a document matrix, in the same way word embeddings are assembled into a sentence matrix at the sentence level.

\subsection{Convolution}

A convolutional layer contains a filter bank $\vF \in \R^{d\times w_f \times n_f}$ where $w_f$ and $n_f$ refer to the width and number of feature maps respectively.  The first dimension of each feature map $\vf \in \R^{d\times w_f}$ is equal to the number of dimensions in the embeddings generated by the layer below.

The convolution operation in our model is one dimensional.  We align the first axis of each feature map with the embedding axis and convolve along the rows.  At the sentence level this corresponds to convolving across words and at the document level it corresponds to convolving across sentences.

Each feature map generates a 1d row of numbers, where each value is obtained by applying the feature map at a different location along the sentence matrix.  The outputs of different feature maps are then stacked to form a new matrix of latent representations which is fed as input to the next layer.  In all cases we use wide (``\texttt{full}'') convolutions in order for all weights in the feature maps to reach every word/sentence, including ones on the edges.

\begin{figure}
\centering
\includegraphics[width=0.6\linewidth]{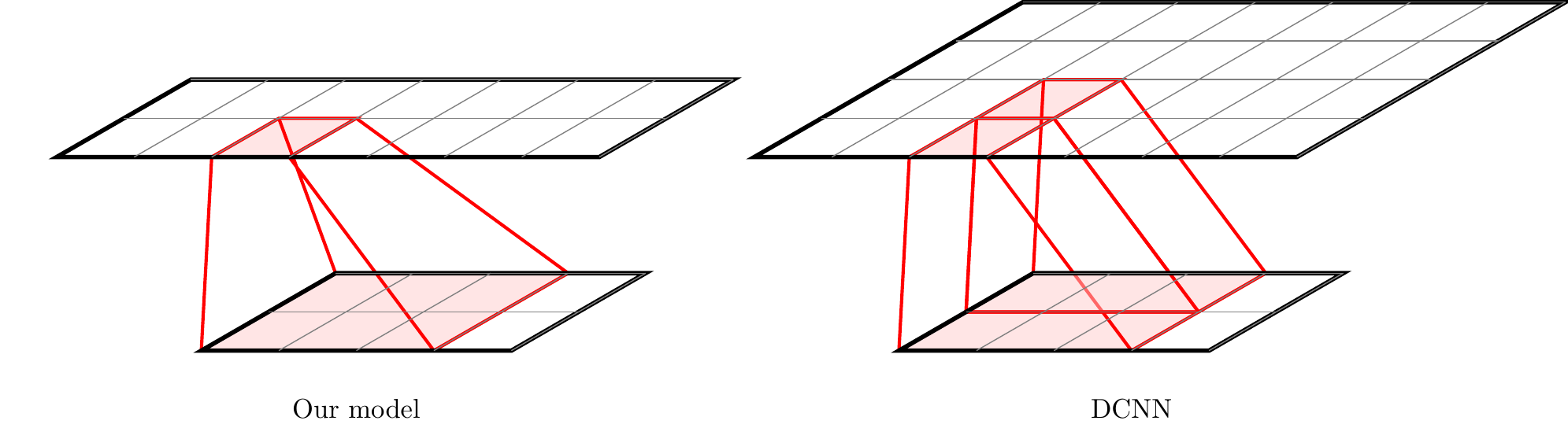}
\includegraphics[width=0.37\linewidth]{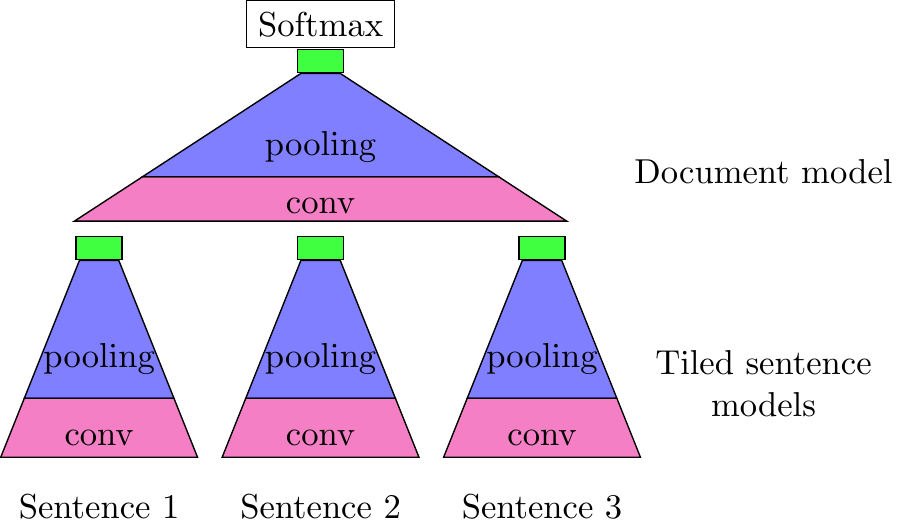}
\caption{\textbf{Left:} A comparison between the convolutions in our model (left) and in the model of Kalchbrenner \emph{et~al.}~\cite{NalKalchbrennerGrefenstette2014} (right), both using two feature maps (only one is shown).  In our model each feature map looks at every dimension of the layer below and generates a single value for each position along the sentence matrix, in the model of Kalchbrenner \emph{et~al.}\ each row of each feature map generates an output in the layer above, so two feature maps convolved with embeddings of dimension two generate four outputs. \textbf{Right:} A schematic of our full model.  Only one layer of convolution and pooling is shown in each level, but any part of the model can be made deeper.}
\label{fig:compare-conv-style}
\end{figure}

Unlike the DCNN, we treat embedding dimensions as channels.  In the DCNN each dimension of each feature map generates its own hidden representation, i.e.\ a $d \times w_f$ feature map generates a representation of size $d \times |s| + w_f - 1$.  In our model the same feature map generates a representation of size $1 \times |s| + w_f - 1$.  Our approach also obviates the need for the ``folding'' operation that appears in the DCNN.  This difference is illustrated in Figure~\ref{fig:compare-conv-style} (left).  This change also means our model has substantially fewer parameters than the DCNN, since the output of each convolution layer is smaller by a factor of $d$.

Our approach corresponds to the typical setup in computer vision where each feature map is small in the spatial domain, but spans all channels of the input image~\cite{Krizhevsky2012}.  Since we work with text, there is only one ``spatial'' dimension (the direction of reading) and the different embedding dimensions correspond to channels.

\subsection{$k$-max pooling}

Since different sentences and documents have different lengths, not all embedding matrices will be of the same width.  This is not an issue for the convolutional layers, since convolutions can handle inputs of arbitrary width, but is problematic to use as input to a fully connected layer, or if we want to generate sentence embeddings for use by another model which expects fixed size inputs.

The solution to this is $k$-max pooling \cite{NalKalchbrennerGrefenstette2014}, which is applied to each row of the embedding matrix separately.  To apply $k$-max pooling to a single row, we keep the $k$ largest values along that row and discard the rest.  Since $k$ is a fixed parameter this always generates a fixed size output (if the input has length less than $k$ we pad it with zeros).  For example, applying 2-max pooling to $[3,1,5,2]$ yields $[3,5]$.  This procedure is also illustrated graphically in Figure~\ref{fig:cdm}.

It is not necessary to impose a fixed length on the representations if the next layer is a convolution layer; however, some pooling is desirable to decrease the size of the representation and has proved to be very effective in computer vision.  The compromise proposed by Kalchbrenner \emph{et~al.}~called \emph{dynamic $k$-max pooling} is to set $k$ to be a fraction of the length of the input sentence.  In this way the pooling operation discards (say) half of the inputs instead of all-but-$k$ of them, which can help information from long sentences or documents to propagate through the model.






\section{Application to multiple tasks}

Learning a single model that is capable of solving multiple tasks has been one of the holy grails of the field of machine learning. Our ConvNet approach is strongly motivated by this vision. In this section, we will demonstrate that we can learn a single ConvNet model to visualise the saliency of words and sentences in documents, to classify sentences and documents and to summarise documents. 

\subsection{Classification}

In the preceding section, we argued that our ConvNet model has substantially fewer parameters than the ConvNet model of Kalchbrenner~\emph{et~al.}~\cite{NalKalchbrennerGrefenstette2014}.  Parameter parsimony is an important feature for scaling up our models or embedding them in mobile devices. 
It is natural to ask ourselves if we have paid a price in classification performance to attain this feature. 

Our first experiment, which closely reproduces the tweet sentiment classification setting of Kalchbrenner~\emph{et.~al.}~\cite{NalKalchbrennerGrefenstette2014}, shows that both models achieve comparable results.  The training set contains 1.6 million tweets with weak sentiment labels that were automatically derived from the presence of emoticons in the tweet text~\cite{Go2009}.  The test set contains 359 tweets where the sentiment label has been assigned by a human annotator.

The model we use has two layers of convolution, pooling and tanh transformations using 6 and 14 feature maps with widths 7 and 5, respectively. The word embeddings are 60 dimensional.  This exactly follows the set up used in Kalchbrenner~\emph{et.~al.}~\cite{NalKalchbrennerGrefenstette2014}, except that we have used our own version of the convolution operation and our model has no sum pooling. As a consequence, our model has substantially fewer parameters than the model of Kalchbrenner~\emph{et.~al.}.
The results are reported in Table~\ref{tab:perf-tables}, which also lists a selection of results of previous work on this data set for comparison. Both our model, with 46 errors, and Kalchbrenner's, with 45 errors, are very close in performance and significantly better than other competitors.

\begin{table}
\begin{center}
\small{
\begin{tabular}{l r}
\hline
Model     & Errors \\ \hline\hline
SVM       & 66 \\ 
BiNB      & 62 \\ 
MaxEnt    & 61 \\ \hline
Max-TDNN  & 76 \\ 
NBoW      & 68 \\ 
DCNN       & 45 \\ \hline
Our Model & 46 \\ \hline
\end{tabular}
\qquad\qquad
\begin{tabular}{lr}
\hline
Model & Accuracy \\ \hline\hline
BoW (b$\Delta$t'c)  & 88.23\% \\
Full+BoW            & 88.33\% \\ 
Full+Unlabelled+BoW & 88.89\% \\ \hline
WRRBM               & 87.42\% \\ 
WRRBM+BoW (bnc)     & 89.23\% \\ \hline
SVM-bi              & 86.95\% \\
NBSVM-uni           & 88.29\% \\
NBSVM-bi            & 91.22\% \\ \hline
Paragraph Vector    & 92.58\% \\ \hline
Our model           & 89.38\% \\ \hline
\end{tabular}
}
\end{center}
\caption{\textbf{Left:} Number of test set errors on the twitter sentiment dataset.  The first block of three entries is from Go \emph{et~al.}~\cite{Go2009}, the second block is from Kalchbrenner \emph{et~al.}~\cite{NalKalchbrennerGrefenstette2014}. \textbf{Right:} Error rates on the IMDB movie review data set.  The first block is from Maas~\emph{et~al.}~\cite{Maas2011}, the second from Dahl~\emph{et~al.}~\cite{Dahl2012}, the third from Wang and Manning~\cite{Wang2012a} and the fourth from Le and Mikolov~\cite{Le2014}.}
\label{tab:perf-tables}
\end{table}

A central goal of this paper is to develop novel ConvNet methods for visualisation and summarisation of reviews. However, to do well in these tasks we also want the same ConvNet to work well as a review sentiment classifier. This is also in line with our aspiration of building models that can be deployed to solve multiple tasks. 

The previous classification experiment showed that our model is a very good sentence classifier. In the next experiment, we show that it is also a good document classifier.

We focus on the IMDB movie review sentiment data set, which was originally introduced by Maas~\emph{et~al.}~\cite{Maas2011} as a benchmark for sentiment analysis.  The dataset contains a total of 100000 movie reviews posted on IMDB.  There are 50000 unlabelled reviews and the remaining 50000 reviews are divided into a 25000 review training set and a 25000 review test set.  Each of the labelled reviews has a binary label, either positive or negative.  In our experiments, we train only on the labelled part of the training set.

We pre-process each review by first stripping HTML markup and breaking the review into sentences and then breaking each sentence into words.  We use NLTK\footnote{\url{http://www.nltk.org/}} to perform these tasks.  We also map numbers to a generic \texttt{NUMBER} token, any symbol that is not in \texttt{.?!} to \texttt{SYMBOL} and any word that appears fewer than 5 times in the training set to \texttt{UNKNOWN}.  This leaves us with a 29493 word vocabulary.

Our best model for this data set uses a one layer ConvNet model to process each sentence, followed by a one layer ConvNet model to process each document.  The word embeddings are 10 dimensional and the sentence model uses 6 feature maps of width 5 followed by a $k$-max pooling layer of width 4, which leads to sentence embeddings with 360 dimensions.  The document model uses 15 feature maps that each look at 5 adjacent sentences followed by a $k$-max pooling layer of width 2 which leads to document embeddings with 30 dimensions.

The results of this experiment are shown in Table~\ref{tab:perf-tables}.  Our model achieves what is, to the best of our knowledge, the third best published result on this data set. We found this result to be very encouraging because this dataset is too small to train a ConvNet model as well as we would like too.  Our main challenge in achieving good performance on this task was regularising the model strongly enough that it would not overfit.

The best result on this data set is achieved by the paragraph vector of Le and Mikolov~\cite{Le2014} .  However, inference in their model is expensive, since they must perform an optimisation in order to infer the paragraph vector for an unseen document.  In contrast, we are able to compute a document embedding using a single feed forward pass.

Having shown that we can train our ConvNet model to be a good sentiment classifier, in the following section we capitalise on this to show that our model also enables us to visualise salient features of documents and to provide users with compact summaries of reviews.

\subsection{Visualisation and Summarisation}

In this section we show that recent work in visualising the activations of ConvNets in computer vision can also be applied to visualising ConvNets for text.  In addition to providing insights into what the model has learned, the same techniques can be used to extract automatic summaries of texts.

Deconvolutional networks~\cite{Zeiler2010,Zeiler2011} have been used to great effect to generate interpretable visualisations of the activations in deep layers of convolutional neural networks used in computer vision~\cite{Zeiler2012}.
More recent work has shown that good visualisations can be obtained by using a single backpropogation pass through the network.  In fact this procedure is formally quite similar to the operations carried out in a deconvolutional net~\cite{Simonyan2013}.  Visualisation through backpropogation is a generalisation of the deconvolutional approach, since one can backpropogate through non-convolutional layers.

The first step in our summarisation procedure is to create a saliency map for the document by assigning an importance score to each sentence.  To generate the saliency map for a given document, we adopt the technique of  Simoyan~\emph{et~al.}~\cite{Simonyan2013} with a modified objective function.

We first perform a forward pass through the network to generate a class prediction for the document.  We then construct a \emph{pseudo-label} by inverting the network predictions, and feed this to the training loss function as the true label.  This choice of pseudo-label allows us to induce the greatest loss.

To infer saliency for words we take a first order Taylor expansion of the loss function using the pseudo-label.  Formally we use the network function $f(x)$, where $x$ denotes the words that compose the document being summarised.  We approximate the loss as a linear function of $x$
\begin{align*}
L(\tilde{y}, f(x)) \approx w^T x + b
\end{align*}
where $\tilde{y}$ is the inverted label and
\begin{align*}
w = \frac{\partial L}{\partial x}\bigg|_{(\tilde{y}, f(x))} \enspace.
\end{align*}
The vector $w$ has one entry for each word in the document and we can use $|w_i|$ as a measure of the saliency of word $i$.  These saliency scores can be easily computed by performing a single pass of backpropogation through the network.  The intuition behind using gradient magnitudes as a saliency measure is that the magnitude of the derivative indicates which words need to be changed the least to affect the score the most.

We have explained how to generate saliency maps for words but we can use the same technique to generate saliency maps for sentences in the same way, as our model has a clear separation between sentence and document level representations.  To generate sentence level saliency we simply perform a partial backpropogation pass through the model, or equivalently we take the Taylor expansion with respect to a partial evaluation of the network.  That is,
\begin{align*}
w = \left( \frac{\partial L}{\partial f_s(x)}\bigg|_{(\tilde{y}, f(x))} \right)^T f_s(x)
\end{align*}
where $f_s$ denotes the evaluation of the network up to the sentence level.

We use the saliency scores for each sentence to rank the sentences in each document.  To generate a summary of a fixed length $k$ we simply take the $k$ most highly ranked sentences from the review and use them to form the summary.

\subsubsection{Evaluation}

In order to evaluate the automatic summaries produced by our model we train a Na\"ive Bayes classifier on the IMDB movie review sentiment data set and use it to classify our review summaries. We use TF/IDF weighted unigram features with no further processing to train the Na\"ive Bayes model.

The results of this experiment are shown in Table~\ref{tab:naive-bayes}.  We compare the accuracy of Na\"ive Bayes to summaries of different sizes created by taking the top ranked sentences using the visualisation technique.  Even keeping only 20\% of each review the accuracy of the classifier trained on full reviews drops by less than 1\% on the test set.  As a baseline we also report the accuracy of the same classifier on summaries created by choosing random sentences from each review, and it is clear from the results that the summaries we create preserve a significant amount of information that is lost by choosing random sentences.

We also compare against the common summarisation heuristic of building a summary by choosing the first and last sentence of each review.  This heuristic performs particularly badly on this data set, which can be explained by the the fact that many reviews begin with a few sentences of plot summary, which are generally not relevant to the sentiment of the review.

We show several examples of summaries created by our model in Figure~\ref{fig:example-summaries}.  As can be seen from the examples, many of the reviews begin with short descriptions of where the reviewer saw the film, or with a brief summary of the plot.  These sentences are not useful as part of the summaries since they do not express an opinion on the movie being reviewed.  Our model learns to ignore these background sentences very consistently.

\begin{table}
\begin{center}
\begin{tabular}{lrrr|lrrr}
Proportion & Summary & Random & Margin & Fixed & Summary & Random & Margin \\ \hline\hline
100\% & 83.03 & 83.03 &   --- &      ~ & ~& ~\\
 50\% & 83.53 & 79.79 & +3.74 & Pick 5 & 83.07 & 80.02 & +3.05 \\
 33\% & 83.10 & 76.72 & +6.38 & Pick 4 & 83.09 & 79.05 & +4.04 \\
 25\% & 82.91 & 74.87 & +8.04 & Pick 3 & 82.88 & 77.15 & +5.73 \\
 20\% & 82.67 & 73.20 & +9.47 & Pick 2 & 82.04 & 74.48 & +7.56 \\ \hline
First and last & 68.62
\end{tabular}
\end{center}
\caption{Results of classifying summaries with Na\"ive Bayes.  Results labelled proportion indicate selecting up to the indicated percentage of sentences in the review, and results labelled fixed show the result of selecting a fixed number of sentences from each.  The summary column shows the accuracy of Na\"ive Bayes on summaries produced by our model.  The random column shows the same model classifying summaries created by selecting sentences at random.  The margin column shows the difference in accuracy between our model and the random summaries.}
\label{tab:naive-bayes}
\end{table}




\begin{figure}[t]
\tiny{I caught this movie on the Sci-Fi channel recently. It actually turned out to be pretty decent as far as B-list horror/suspense films go. \highlight{Two guys (one naive and one loud mouthed a **) take a road trip to stop a wedding but have the worst possible luck when a maniac in a freaky, make-shift tank/truck hybrid decides to play cat-and-mouse with them.} Things are further complicated when they pick up a ridiculously whorish hitchhiker. What makes this film unique is that the combination of comedy and terror actually work in this movie, unlike so many others. The two guys are likable enough and there are some good chase/suspense scenes. Nice pacing and comic timing make this movie more than passable for the horror/slasher buff. \highlight{Definitely worth checking out.}}
\\[0.2cm]
\tiny{I just saw this on a local independent station in the New York City area. \highlight{The cast showed promise but when I saw the director, George Cosmotos, I became suspicious. And sure enough, it was every bit as bad, every bit as pointless and stupid as every George Cosmotos movie I ever saw.} He's like a stupid man's Michael Bey -- with all the awfulness that accolade promises. There's no point to the conspiracy, no burning issues that urge the conspirators on. We are left to ourselves to connect the dots from one bit of graffiti on various walls in the film to the next. Thus, the current budget crisis, the war in Iraq, Islamic extremism, the fate of social security, 47 million Americans without health care, stagnating wages, and the death of the middle class are all subsumed by the sheer terror of graffiti. A truly, stunningly idiotic film.}
\\[0.2cm]
\tiny{Graphics is far from the best part of the game. \highlight{This is the number one best TH game in the series.} Next to Underground. \highlight{It deserves strong love. It is an insane game.} There are massive levels, massive unlockable characters... it's just a massive game. \highlight{Waste your money on this game. This is the kind of money that is wasted properly.} And even though graphics suck, thats doesn't make a game good. Actually, the graphics were good at the time. Today the graphics are crap. WHO CARES? As they say in Canada, This is the fun game, aye. (You get to go to Canada in THPS3) Well, I don't know if they say that, but they might. who knows. Well, Canadian people do. Wait a minute, I'm getting off topic. This game rocks. Buy it, play it, enjoy it, love it. It's PURE BRILLIANCE.}
\\[0.2cm]
\tiny{The first was good and original. I was a not bad horror/comedy movie. So I heard a second one was made and I had to watch it . What really makes this movie work is Judd Nelson's character and the sometimes clever script. \highlight{A pretty good script for a person who wrote the Final Destination films and the direction was okay.} Sometimes there's scenes where it looks like it was filmed using a home video camera with a grainy - look. Great made - for - TV movie. \highlight{It was worth the rental and probably worth buying just to get that nice eerie feeling and watch Judd Nelson's Stanley doing what he does best.} I suggest newcomers to watch the first one before watching the sequel, just so you'll have an idea what Stanley is like and get a little history background.}
\\[0.2cm]
\tiny{When the movie was released it was the biggest hit and it soon became the Blockbuster. But honestly the movie is a ridiculous watch with a plot which glorifies a loser. The movie has a Tag - line - ``Preeti Madhura, Tyaga Amara'' which means Love's Sweet but Sacrifice is Immortal. \highlight{In the movie the hero of the movie (Ganesh) sacrifices his love for the leading lady (Pooja Gandhi) even though the two loved each other!} His justification is the meaning of the tag - line. This movie influenced so many young broken hearts that they found this ``Loser - like Sacrificial'' attitude very thoughtful and hence became the cult movie it is, when they could have moved on with their lives. \highlight{Ganesh's acting in the movie is Amateurish, Crass and Childishly stupid.} He actually looks funny in a song, (Onde Ondu Sari ...) when he's supposed to look all stylish and cool. His looks don't help the leading role either. \highlight{His hair style is badly done in most part of the movie.} POOJA GANDHI CANT ACT. Her costumes are horrendous in the movie and very inconsistent. \highlight{The good part about the movie is the excellent cinematography and brilliant music by Mano Murthy which are actually the true saving graces of the movie.} Also the lyrics by Jayant Kaikini are very well penned. The Director Yograj Bhat has to be lauded picturization the songs in a tasteful manner. Anyway all - in - all except for the songs, the movie is a very ordinary one !!!!!!}
\\[0.2cm]
\tiny{A friend and I went through a phase some (alot of) years ago of selecting the crappest horror films in the video shop for an evening's entertainment. For some reason, I ended up buying this one (probably v. v. cheap). \highlight{The cheap synth soundtrack is a classic of its time and genre.} There's also a few very amusing scenes. Among them is a scene where a man's being attacked and defends himself with a number of unlikely objects, it made me laugh at the time (doesn't seem quite so funny in retrospect but there you go). \highlight{Apart from that it's total crap, mind you.} But probably worth a watch if you like films like ``Chopping Mall''. Yes, I've seen that too.}
\\[0.2cm]
\tiny{I tried restarting the movie twice. I put it in three machines to see what was wrong . Did Steven Seagal's voice change? \highlight{Did he die during filming and the studio have to dub the sound with someone who doesn't even resemble him?} Or was the sound on the DVD destroyed? After about 10 minutes, you finally hear the actor's real voice. Though throughout most of the film, it sounds like the audio was recorded in a bathroom. \highlight{I would be ashamed to donate a copy of this movie to Goodwill, if I owned a copy.} I rented it, but I will never do that again. I will check this database before renting any more of his movies, all of which were (more or less) good movies. \highlight{You usually knew what you were getting when you watched a Steven Seagal movie.} I guess that is no more.}
\\[0.2cm]
\tiny{Vertigo co - stars Stewart (in his last turn as a romantic lead) and Novak elevate this, Stewart's other ``Christmas movie,'' movie to above mid - level entertainment. \highlight{The chemistry between the two stars makes for a fairly moving experience and further revelation can be gleaned from the movie if witchcraft is seen as a metaphor for the private pain that hampers many people's relationships.} All in all, a nice diversion with legendary stars, 7/10}
\\[0.2cm]
\caption{Several example summaries created by our ConvNet.  The full text of the review is shown in black and the sentences selected by the ConvNet appear in colour. While summarising a review with the first sentence is a popular pragmatic approach, it is clear in these examples that this heuristic is not as effective as the ConvNet summarisation scheme.  Each summary is created by selecting up to 20\% of the sentences in the review.}
\label{fig:example-summaries}
\end{figure}


\section{Conclusion}

In this paper we have introduced a convolutional neural network model that is able to represent the meaning of documents by embedding them in a low dimensional vector space while preserving distinctions of word and sentence order, crucial for capturing nuanced semantics.

Our model builds the document representation in a compositional manner by combining word embeddings into sentence embeddings and then further combining sentence embeddings into a representation for the full document.

We have shown that a single model can be used to accomplish a wide variety of document modelling tasks, including classification and summarisation and visualisation of document structure.  These tasks are all accomplished with a single model trained once on a classification task and no re-training is needed to apply the model beyond the task for which it was created.

The structure of our model allows us to learn word, sentence and document representations simultaneously.  An important avenue for future work work is exploring how representations at all three levels can be further exploited.  Another important directions for future work is understanding how our model can be trained using unlabelled data.

\clearpage
\small{
\bibliography{txtnets}
\bibliographystyle{abbrv}
}

\end{document}